\documentclass[letterpaper]{article} % DO NOT CHANGE THIS
\usepackage{aaai2026}  % DO NOT CHANGE THIS
\usepackage{times}  % DO NOT CHANGE THIS
\usepackage{helvet}  % DO NOT CHANGE THIS
\usepackage{courier}  % DO NOT CHANGE THIS
\usepackage[hyphens]{url}  % DO NOT CHANGE THIS
\usepackage{graphicx} % DO NOT CHANGE THIS
\usepackage{natbib}  % DO NOT CHANGE THIS AND DO NOT ADD ANY OPTIONS TO IT
\usepackage{caption} % DO NOT CHANGE THIS AND DO NOT ADD ANY OPTIONS TO IT
\usepackage{algorithm}
\usepackage{algorithmic}

\usepackage{newfloat}
\usepackage{listings}
\DeclareCaptionStyle{ruled}{labelfont=normalfont,labelsep=colon,strut=off} % DO NOT CHANGE THIS
\floatstyle{ruled}
\newfloat{listing}{tb}{lst}{}
\floatname{listing}{Listing}
\usepackage{xcolor}
\newcommand{\answerYes}[1]{\textcolor{blue}{#1}} 
 
\newcommand{\answerNA}[1]{\textcolor{gray}{#1}}

\nocopyright %-- Your paper will not be published if you use this command
\usepackage{fancyhdr}
\fancypagestyle{empty}{%
  \fancyhf{} 
  \fancyfoot[C]{\textit{This paper has been accepted for the upcoming 20th International AAAI Conference on Web and Social Media (ICWSM 2026)}}
  
}

\title{Ideology Prediction of German Political Texts}
\author {
    Sinclair Schneider\textsuperscript{\rm 1},
    Florian Steuber\textsuperscript{\rm 1},
    João A. G. Schneider\textsuperscript{\rm 1}, % ,\rm 2
    Gabi Dreo Rodosek\textsuperscript{\rm 1}
}
\affiliations {
    \textsuperscript{\rm 1}Bundeswehr University Munich\\
    Research Institute CODE\\
    Carl-Wery-Straße 22\\
    81739, Munich, Bavaria, Germany\\
    sinclair.schneider@unibw.de, florian.steuber@unibw.de, joao.schneider@unibw.de, gabi.dreo@unibw.de
}

\usepackage{bibentry}
\begin{document}

\maketitle

\begin{abstract} %Do not include references in your abstract.
% {\color{red}
Elections represent a crucial milestone in a nation's ongoing development. To better understand the political rhetoric from various movements, ranging from left to right, we propose a transformer-based model capable of projecting the political orientation of a text on a continuous left-to-right spectrum, represented by a normalized scalar, $d \in \left[-1, 1\right]$.
This approach enables analysts to focus on specific segments of the political landscape, such as conservatives, while excluding liberal and far-right movements. Such a task can only be achieved with multiclass classifiers, provided that the desired orientation is incorporated within one of their predefined classes.
To determine the most suitable foundation model among 13 candidate transformers for this task, we constructed four distinct corpora. One corpus comprised annotated plenary notes from the German Bundestag, while another was based on an official online decision-making tool, Wahl-O-Mat. 
The third corpus consisted of articles from 33 newspapers, each identified by its political orientation, and the fourth included 535,200 tweets from 597 members of the 20th and 21st German Bundestag.
To mitigate overfitting, we used two distinct corpora for training and two for testing, respectively. For in-domain performance, DeBERTa-large achieved the highest F1 score ($F_1=0.844$) as well as for the X (Twitter) out-of-domain test ($ACC=0.864$). Regarding the newspaper out-of-domain test, Gemma2-2B excelled ($MAE = 0.172$). 
This study demonstrates that transformer models can recognize political framing in German news at the level of public opinion polls. 
Our findings suggest that both the model architecture and the availability of domain-specific training data can be as influential as model size for estimating political bias. 
We discuss methodological limitations and outline directions for improving the robustness of bias measurement.

\end{abstract}

% Uncomment the following to link to your code, datasets, an extended version or similar.
% You must keep this block between (not within) the abstract and the main body of the paper.
\begin{links}
    \link{Code}{https://github.com/SinclairSchneider/german_ideology_prediction}
    \link{Bundestag/Wahl-O-Mat Datasets}{https://doi.org/10.57967/hf/4924}
    \link{German Media Datasets}{https://huggingface.co/collections/SinclairSchneider/german-media-67dcb6c0bf4c007db3999153}
    %\link{Extended version}{https://aaai.org/example/extended-version}
\end{links}

\section{Introduction} % ca. 1.5 Seiten
% \section{Related Work}
%\subsubsection{Problem}
% {\color{red}
In February 2023, investigative journalists from the network ``Forbidden Stories'' uncovered a disinformation-as-a-service provider, working with social media bot accounts, known as ``Team Jorge'' \cite{andrzejewski_team_2023}. This entity claims to have manipulated 33 elections, 27 of which were deemed successful. To demonstrate their capabilities, Team Jorge spread false rumors about a deceased emu (\#RIP\_Emmanuel), which ultimately led to real issues at the animal's farm. Although this is a particularly negative example, it highlights the considerable influence of social media on politics.

%paragraph
We believe that the robust tools of social media analysis can play a valuable role in helping political parties better understand the needs and preferences of their constituents, as well as in forecasting the trajectory of political discourse. To achieve this goal, the political ideology spectrum can be quantified on a continuous scale from -1 (left) to 1 (right).
Assuming such a mapping is found, individuals' political ideology can be approximated from tweets on X. 
A range of $-1 \leq \theta \leq -0.9$ would yield left-wing topics such as the establishment of a single public healthcare system, the withdrawal of U.S. troops from Germany, a focus on social justice and climate protection, and an end to weapons exports. More %liberal
centrist positions may be found in a range of $-0.1 \leq \theta \leq 0.1$, including principles against extremism, efforts to combat hate speech and misinformation, democratic values, military modernization, and digital strategies. Consequently, a threshold of $0.9 \leq \theta \leq 1$ might reveal right-wing topics such as the end of weapon supplies to Ukraine, claims of economic destruction linked to voting for the Green Party, viewing climate change as a business model, and the perception of immigration and Islam as threats to Western countries.

To achieve this, one could implement a topic modeling algorithm such as BERTopic \cite{grootendorst_bertopic_2022}. However, these approaches lack an essential component: the ability to dynamically focus on a specific political direction, which can only be addressed partially by classifiers with predefined categories. Therefore, this paper introduces a new algorithm that maps political texts onto a continuous scale ranging from -1 to 1, with a liberal orientation at 0.

This paper addresses three significant challenges: first, it aims to map text onto a continuous left-to-right spectrum rather than simply categorizing it into discrete classes. Second, it seeks to adapt the generated algorithm to account for local political biases through a semi-supervised labeling approach. Third, it focuses on ensuring the algorithm’s effectiveness by testing on distinct, out-of-domain datasets.

\subsubsection{Approach}

The foundation for training a classifier that maps texts to a continuous left-to-right spectrum is the association of two-dimensional normalized vectors with political parties. An entirely left-wing party would be represented by a vector pointing to the left (-1, 0), while a right-wing party would have a vector directed to the right (1, 0). A centrist party would be indicated by an upward vector towards the center (0, 1). Intermediate positions are encoded by vectors of unit length at corresponding angles. 

The output of a trained multilabel classifier, indicating the extent to which a party agrees with a given statement, is then multiplied by the corresponding vectors. At the end, all vectors are added, and the angle of the newly formed vector represents the classification result.
To demonstrate that this approach is effective, it is finally tested on both crawled German newspapers and politicians' tweets, for which the political leanings are known.  
This outlines both the classifier's accuracy and its out-of-domain capabilities.
In order to do so, we trained and tested 13 transformer classifiers.

\subsubsection{Contribution}
The main contributions of this paper are the extension of previous approaches that used categorical variables with a continuous left-right spectrum between -1 and 1, as well as %the proof of
demonstrating the out-of-sample capabilities of our classifier. When tested against the 33 newspapers, our best classifier yielded a mean error (ME) of 0.17 on a scale between -1 and 1, which is an error of 8.58\% on a survey-based benchmark dataset. Regarding the origin-prediction tweets, we found that accuracy increases to 0.864 when 100+ words are available. By using plenary speeches from the German Bundestag as one of the training sets, we ensured that our classifier is perfectly aligned with the German left-right spectrum without introducing the author's bias. With a total of four self-collected datasets, we also made sure that the out-of-domain accuracy is provided. 
By adapting the task of political stance prediction to a German context, we contribute to a more diverse array of training data and models, as this not only requires linguistic adaptation but also considers the unique political environment.

\section{Related Work}
Political ideology detection is typically done by building classes such as left, center, or right, using a manual annotation approach \cite{baly-etal-2020-detect}. 

Different research projects approach the issue of such a limited political scale in various ways. Some focus solely on detecting (extreme) left-wing or right-wing opinions \cite{kiesel-etal-2019-semeval, jakob_augmented_2024}, while others offer a broader spectrum \cite{allsides-website}. These broader approaches include classifications for ``lean left'' and ``lean right'', situated between the center and the two extremes.
Others offer an even more fine-grained classification of seven or more classes \cite{preotiuc-pietro-etal-2017-beyond, fagni_fine-grained_2022}, for instance, very conservative, conservative, moderately conservative. 

Most foundational research is conducted in English, which often leads to an association with the United States. However, simply translating existing English-language datasets is insufficient for their application to German politics, given the diverse political views across countries. For this reason, researchers have begun to collect and label specific datasets in German, utilizing information from German newspapers \cite{aksenov_fine-grained_2021}.

The global nature of social media platforms, which span across borders and cultures, makes it difficult to develop generalizable models trained on tweets. For instance, methods that achieve over 90\% accuracy on a carefully selected dataset can drop to approximately 65\% when applied to different users within the same network \cite{cohen_classifying_2013}. Despite this, social media continues to be a focal point for transformer-based classification methods, particularly with models tailored for social media like BERTweet \cite{nguyen-etal-2020-bertweet} and PoliBERTweet \cite{kawintiranon-singh-2022-polibertweet}.

Expanding beyond a text-only approach to ideology classification and incorporating users' networks opens up new opportunities for classification methods that utilize transformers, as demonstrated in previous research \cite{RetweetBERT}. 

Exploring publications analyzing German Bundestag speeches leads us to the work of Erhard et al. \shortcite{PopBERT}, who investigated the rise of populism using these speeches. They identified four main categories: anti-elitism, people-centrism, left-wing ideology, and right-wing ideology. This framework enhances the traditional two-dimensional political spectrum by incorporating anti-elitism and people-centrism, while still relying on hand-labeled discrete categories.

Baly et al. \shortcite{baly-etal-2019-multi} adopt a similar approach by introducing trustworthiness as a second dimension on a three-point scale. Their work demonstrates that political orientation can be a useful factor in detecting misinformation, bias, and propaganda.

The issue of models trained on specific domains, such as news sites, performing poorly on other domains, like social media, in ideology classification has been noted by \citeauthor{volf_political_2025} \shortcite{volf_political_2025}. They addressed this challenge by mixing datasets from multiple domains for the training process. 
Another way to improve the classifier's output is to build a dataset comprising the same stories told by news outlets with different political biases, providing a direct comparison of the same story across different political perspectives \cite{liu-etal-2022-politics}.

All approaches discussed so far are limited due to their categorical outputs. Specifically, ordinal scales cannot measure the extent to which left- or right-leaning perspectives are present. As there is no convention regarding the specific categories, model usage is limited to a predefined context. For instance, the concept of a left-wing opinion in the US may differ significantly from that in Germany.

\section{Methodology}
    The processing pipeline was structured as follows: First, data from several sources was collected and further enriched to obtain generalizable models. Second, a binary political classifier and subsequent multi-label party classifiers were trained, using multiple BERT, Llama, and Gemma LLMs. Third, the multilabel output was converted to a continuous left-right spectrum (-1 to 1). Finally, in-domain and out-of-domain performance was evaluated using separate test sets, each drawn from an independent dataset. Furthermore, pre- and post-vector-optimization results are compared. 

\subsection{Datasets}

Two independent sources (Bundestag, Wahlomat) were preprocessed for model training and testing. Despite artificially enriching and splitting the data (80:20 train-test split), models may %tend to 
overfit. This is why two additional datasets (newspapers, tweets) were used for model evaluation. For training and evaluation, the data of all datasets were either pre- or auto-labeled as explained below. 

\subsubsection{Bundestag Dataset}

All plenary debates of the German Bundestag are recorded in writing by stenographers and published \cite{bundestag}. Besides the text of the speech, the speaker's name and party membership are minuted. This is also true regarding requests (question, party and name of the questioner) and all other potential speech interruptions, such as interjections, hissing, applause, etc. (type and party, resp. parties). All protocols were collected and processed for the period from October 2017 to September 2024. The raw speech data comprises 34,174 speeches.  

\paragraph{\textit{Labeling}} The combination of speeches and interruptions constitutes a robust auto-labeling approach. All speeches were filtered for recorded interruptions. Speeches without any interruptions were discarded. For the remaining ones, the sentiment was extracted from the comments. The described extraction process is illustrated in Figure \ref{fig:sentiment-bundestag}. This procedure yielded a dataset of 32,246 annotated statements (i.e., pro or contra opinions of parties). The association between parties based on the extracted sentiment is depicted in Figure \ref{fig:sentiment-correlations} (upper triangle).

\paragraph{\textit{Data Enrichment}}
In order for a classifier to correctly categorise not only political speeches but also political statements in general, the linguistic variance of the statements was artificially increased. For this purpose, a LLama 3.1
model was asked to summarize each text in five different versions: In the words of a child, of a teenager, of an adult, of an eloquent person, or as a social media post (tweet). The expanded dataset consisted of 449,209 statements. It was made publicly available \cite{trainsetpoliticalpartybig} after combining it with the Wahlomat dataset, which is described below.

\subsubsection{Wahlomat Dataset}
The German multi-party system makes it difficult for voters to find the party that represents their interests best. Hence, a digital voters' guide called \textit{Wahl-O-Mat} is released ahead of every federal and state election by the Bundeszentrale für politische Bildung (Federal Agency for Civic Education). It consists of several political statements that the user can agree or disagree with (viz. Fig.~\ref{fig:wahlomat-comb} for an example of the federal election in 2025). For this system to function, the respective party positions (approval, neutral, rejection) were officially surveyed in advance by the Federal Agency.

The used data is available online \cite{qualomat}, comprising 1,751 unique statements regarding the elections between 1998 and 2021.

\begin{figure}%[htbp]
  \centering
  \includegraphics[width=.88\linewidth]{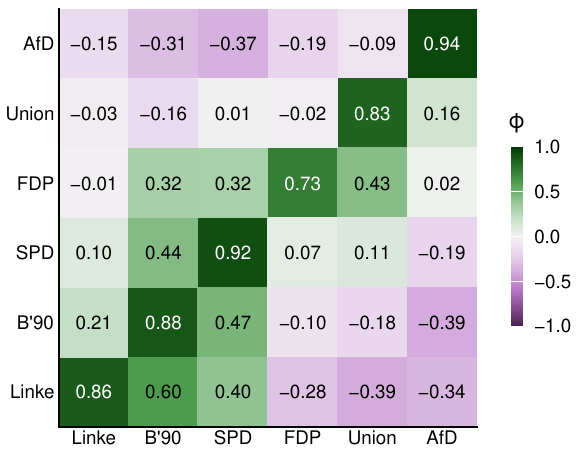}
  \caption{Associations between the parties based on Bundestag sentiments (upper triangle) and Wahlomat statements (lower triangle). Profile similarity (within, viz. diagonal) estimated Pearson's correlation of Phi measures per party between Bundestag and Wahlomat datasets.}
  \label{fig:sentiment-correlations}
\end{figure}

\paragraph{\textit{Labeling}} No annotation was needed as the data already consists of statements and attitudes of all parties. Attitudes were coded as 1 (approval), 0 (neutral), or -1 (rejection), respectively. Based on these values, the association between parties is illustrated in Figure \ref{fig:sentiment-correlations} (lower triangle).

\paragraph{\textit{Data Enrichment}}
The dataset was also synthetically enriched as described above, yielding 87,210 labelled statements. Table \ref{tab:paraphrasing2} presents an example of how the call for introducing a wealth tax could be expressed from various perspectives. The positions of the various parties regarding the original statement and thus also concerning the generated ones can be found in Table \ref{tab:paraphrasing}. 

To ensure that the enriched sentences maintain similarity to the originals, we utilized the Qwen3-Embedding-8B model \cite{zhang_qwen3_2025} to map them into a vector space and calculated the cosine similarity against the original sentences. 
In contrast to parliamentary speeches containing substantial extraneous content (e.g., greetings), the Wahlomat dataset consists exclusively of condensed statements. Hence, only the latter was used for comparisons. 
The overall similarity of the paraphrased examples is 0.74, while the most similar sentences, paraphrased for a teenage audience, yielded an average cosine similarity of 0.78. 
To determine whether political bias was introduced during data enrichment, the cosine similarity distribution is assessed. As is common in statistics, the 5th percentile is computed. Since this extreme quantile is still sufficient with 0.54, we can assume that no fundamental bias has been introduced.

The combined training dataset (Bundestag+Wahlomat) consisted of 570,416 samples and is publicly available \cite{trainsetpoliticalpartybig}.

\subsubsection{Tweet Dataset}
To evaluate the performance of classifiers on short social media texts, we curated a dataset consisting of 535,200 tweets from 597 members of the 20th and 21st German Bundestag (Federal Parliament). Each political party is represented by 89,200 tweets, filtered to include only political content.

\paragraph{\textit{Labeling}}  
The labeling is based on the account owners' affiliation with the respective political party. Each tweet is assigned to a single political party only.

\subsubsection{Newspaper Dataset}

Based on the assumption that the German media landscape sufficiently represents the political spectrum \shortcite[cf. Maurer et al.][]{maurer_fehlt_2024}, a dataset of 33 newspapers was examined. From each source, at least 10,000 articles were collected, resulting in a representative dataset of approximately 10 million articles. An overview with precise numbers for all media is appended (cf. Table \ref{tab:full_medienkompass}). 
Additionally, we retained metadata, such as news categories, to train a binary politics-non-politics classifier that serves as a filter later. 
The dataset was based on prior political classifications available for 39 newspapers (see below). Six newspapers were either discontinued or inaccessible due to technical issues.

\paragraph{\textit{Labeling}}

The political stance of the articles was unknown, but several estimates exist at the newspaper level. The main one used here is based on $n=1148$ participants who rated $k=39$ newspapers on a scale from 1 (\textit{extreme left-wing}) over 4 (\textit{minimal party affiliation}) to 7 (\textit{extreme right-wing}), with fake news and conspiracy theories falling under both extremes, respectively \cite{Medienkompass_2025}.

To verify the validity, we compared the ratings with the ones provided by two independent sources: Firstly, a comparable bias-rating platform that covers various international outlets \cite{mediabiasfactcheck_2025} and secondly, a scientific report about the German media landscape \cite{maurer_fehlt_2024}. Regarding both sources, appropriate association measures were computed using all pairwise complete cases to estimate convergent validity. We also report the respective measures for the subset of our sample.

Mediabiasfactcheck.com reports data for $k=77$ media outlets, but only non-numeric labels in roughly half of the cases. The ratings are based on a scale from -10 (\textit{extrem left}) over 0 (\textit{least biased}) to +10 (\textit{extreme right}). For better comparability, both considered scales were \textit{z}--transformed. Note that this does not affect the correlation estimates but makes the scores directly comparable, as reported in Table \ref{tab:full_medienkompass} (mean values of zero with standard deviations of one). Both estimates were very highly correlated with $r=.90$ (resp. $r=.91$ regarding the sample). However, this estimate was based on the overlap of $k=9$ outlets only ($k'=7$ regarding our sample). To enlarge the intersection, the provided ordinal labels were converted into numerical values (i.e., \textit{left} was assigned to -2, \textit{left-center} to -1, \textit{least biased} to 0, etc. with positive values for the right-hand side). Using Spearman's $\rho$ for ordinal data yielded an even higher correlation of $\rho=.95$ for $k=19$ pairs ($\rho=.96$ for $k'=17$ regarding the sample). 

Although the correlations are very high, it could be criticized that both ratings come from public platforms. Accordingly, the ratings from a scientific study were examined (Maurer et al. \citeyear{maurer_fehlt_2024}), providing data for $k=47$ media outlets by only $n=9$ but extensively trained raters. Here, political ideology was rated using two separate five-point scales. As these showed a strong positive correlation ($r=.63$), both were reduced to a single dimension using principal component analysis (PCA; default settings, varimax rotation). From the resulting one-dimensional values, a subset of $k'=21$ outlets was present at Mediencompass.org, yielding a very high correlation of $r=.95$ ($r=.94$ for the subset of $k'=22$ regarding the sample).

Since ratings were shown to be very highly correlated with two independent sources, the validity of Mediencompass.org can be considered sufficient. This is also the case regarding our sample, which had approximately the same correlation coefficients.

\begin{figure*}[t]
\centering
\includegraphics[width=1\textwidth]{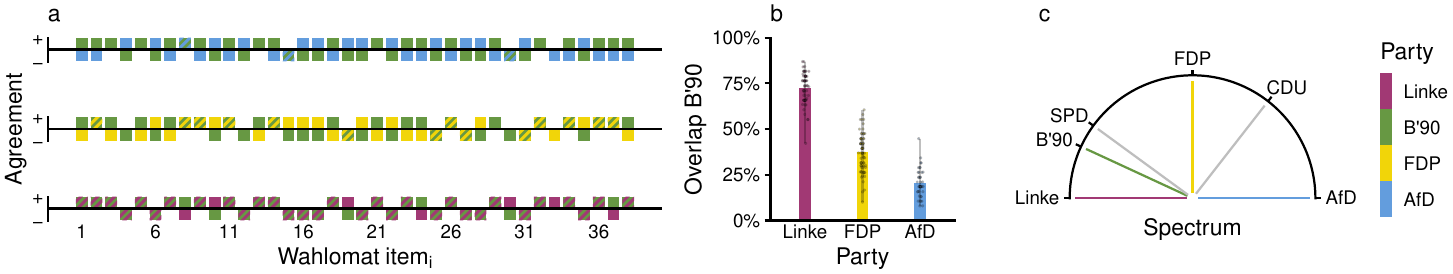} % Reduce the figure size so that it is slightly narrower than the column.
\caption{Exemplary comparison of the Green Party (B'90) against the right-wing (AfD), liberals (FDP), and the left party (Linke) in subplot a. Hachures indicate Wahlomat items on which both compared parties agree or disagree regarding the Brandenburg election in 2024. The mean overlap of all election results is displayed in subplot b. Results are mapped onto a left-right spectrum in subplot c regarding (dis-)similarity distance to the other parties.}
\label{fig:walomat-demonstration}
\end{figure*}

\subsection{Models}
\subsubsection{Foundation Models} \label{sec:foundational_models}

To effectively classify German political texts, we needed to select appropriate foundation models for this multilabel classification task. We used smaller encoder-only models with 0.21-2.1 billion parameters, alongside larger decoder-only models with 1.0-9.0 billion parameters.

For the encoder-only models, we chose DeBERTa Large \cite{dada_impact_2023}, GottBERT Large \cite{scheible_gottbert_2024}, GBERT and GELECTRA Large \cite{chan_germans_2020}, xlm-roberta Large \cite{conneau_unsupervised_2020} and EuroBERT \cite{boizard_eurobert_2025}.
In contrast to the original DeBERTa model presented by \citeauthor{he_deberta_2020}, \citeauthor{dada_impact_2023} trained a model on the same architecture but used a diverse German training corpus. This collection includes online encyclopedias, social media content, legal documents, medical texts, and fiction books, which collectively make the foundation model well-suited for a wide array of German-language applications.

The authors of GottBERT followed a similar approach, except that they employed a RoBERTa BASE architecture \cite{zhuang_robustly_2021} combined with the OSCAR \cite{ortiz_suarez_asynchronous_2019} dataset.

GBERT and GELECTRA Large are developed by the same authors and use the BERT \cite{devlin_bert_2019} and ELECTRA \cite{clark_electra_2020} architectures, respectively, to build German foundation models from German text. The training data for these models is sourced from the OSCAR and OPUS corpora \cite{tiedemann_parallel_2012}, as well as Wikipedia and OpenLegalData \cite{ostendorff_towards_2020}.

The EuroBERT series of models is also an encoder-only architecture trained on 5 trillion tokens across 15 European languages, including German. The family of decoder-only models, including the following, is best known for their generative capabilities but can also be used for classification tasks.

The Gemma 2 models \cite{riviere_gemma_2024} have primarily been trained using English texts, however, they use a significantly larger tokenizer inherited from the Gemini model \cite{rohan_gemini_2025}, comprising 256,000 entries. This extensive vocabulary, combined with the multilingual nature of web data, enables the model to comprehend languages beyond English. A distinctive characteristic of the smaller Gemma models, specifically those with 2 and 9 billion parameters, is that they are trained via knowledge distillation from a larger teacher model. This methodology gives these models an advantage over others of similar size, enhancing their performance and efficacy in various applications.

In contrast, the authors of the Llama 3.2 \cite{dubey_llama_2024} models used FastText to categorize the training data into 176 different languages, including German.

In summary, the smaller encoder-only models are mostly trained on German texts, whereas the larger decoder-only models were trained on multilingual data, including a German component.

\subsubsection{Political Classifier}
Prior to the political orientation classifiers, we require an additional classifier to determine whether a text is political. This is crucial for assessing a newspaper's leaning, whether it is more left or right of center. If we were to classify all texts indiscriminately, we would also include non-political content, which could skew our average classification results toward the center.
For this purpose, we used the metadata described in the previous newspaper section. By merging all political newspaper categories into a single political section and grouping other categories, such as entertainment, separately, we created a well-balanced dataset comprising 234,978 political and non-political texts \cite{sinclair_schneider_trainset_political_text_yes_no_german_nodate}.
This dataset is then used to train a German DeBERTa model \cite{dada_impact_2023} that can predict the probability that a text is politically related. The model is subsequently used with a threshold of 0.8, as suggested by the authors, ensuring that only political texts are processed further. The model achieves an F1 score of 0.99 on the test set, although a slight decline in performance on out-of-domain data is anticipated.

\subsubsection{Political Party Classifiers}
To determine the appropriate political alignment of a text, we have trained 13 classifiers, including DeBERTa-large, EuroBERT, GBERT, XLM-RoBERTa, Llama, and Gemma, using a multilabel classification approach. This approach links an input text to one or more of the six major German political parties. After training, the best-performing classifier among the 13 candidates is evaluated using the out-of-domain newspaper data. The training process itself involves feeding lines similar to those in Table \ref{tab:paraphrasing} into the pretrained foundation models and fine-tuning their weights for four epochs. For a detailed list of models and used parameter specifications, see Table \ref{tab:results}. The six parties Die Linke, Bündnis 90 Die Grünen, SPD, FDP, CDU/CSU, and AfD were selected based on their consistent representation in the German parliament over the past few years. For the training, we focused on likes, excluding dislikes. 

The training was conducted on various GPU servers, ranging from 4 A6000 Ada to 8 H200 GPUs. All training files are publicly accessible \cite{Schneider_german_ideology_prediction}, and the DeBERTA model was executed multiple times to identify the optimal hyperparameter configuration. %, as illustrated in Figure \ref{fig:f1_eval}. 
Given that larger models required several days to train, we were unable to conduct a full training run for each configuration. Instead, we stopped training when the training loss no longer decreased and adjusted the parameters accordingly.

\begin{table}[t]
\centering
\begin{tabular}{lrcr}
  \hline
  Classifier & Size & $F_1$ & train time \\ 
  \hline
  DeBERTa-large$^a$ & 435M & 0.84 & 12:03:20\\
  Gemma-2-9b$^b$ & 9B & 0.82 & 3d 05:04:34\\
  EuroBERT-610m$^b$ & 610M & 0.79 & 1d 06:50:30\\
  Gemma-2-2b$^c$ & 2B & 0.75 & 7d 12:39:57\\
  GottBERT\_large$^b$ & 357M & 0.74 & 09:20:56\\
  gbert-large$^b$ & 337M & 0.73 & 09:32:50\\
  EuroBERT-210m$^b$ & 210M & 0.73 & 09:06:01\\
  gelectra-large$^b$ & 336M & 0.72 & 10:03:12\\
  EuroBERT-2.1B$^b$ & 2.1B & 0.72 & 2d 16:01:31\\
  xlm-roberta-large$^b$ & 561M & 0.71 & 09:37:06\\
  Llama-3.2-3B$^b$ & 3B & 0.71 & 3d 09:40:08\\
  Llama-3.2-1B$^b$ & 1B & 0.69 & 2d 22:43:17\\
  gemma-3-1b$^b$ & 1B & 0.69 & 2d 14:11:47\\
   \hline
\end{tabular}
\caption{Overview of the used models, parameter sizes, evaluation metrics, training hours, and hardware used for training, i.e., \textit{a.} 4 A6000 Ada GPUs (4$\times$48GB vRAM); \textit{b.} 8 H100 GPUs (8$\times$80GB); \textit{c.} 8 H200 GPUs (8$\times$141GB) }
\label{tab:results}
\end{table}

\subsection{From Multilabel to a Continuous Scale}
At this stage, %the reader may question 
it is necessary to explain why we have introduced a multilabel classifier while simultaneously representing a continuous output for political direction on a scale from -1 (left-wing) to 1 (right-wing). The key missing element is an adaptation model that translates the outputs of the multilabel classifiers (which correspond to six political parties) into the left-right spectrum. This adaptation model is based on the premise that each political party can be positioned on a left-right continuum, with varying degrees of liberalism. An alternative geometric representation is a semicircle. In this representation, we position three fixed points: Die Linke, the most left-wing party, on the far left; the FDP, an economically liberal German party, at the center; and the AfD, a strongly right-oriented party, at the far right end of the semicircle.

The remaining task is to determine the positioning of the other three parties. Based on known political positions of the German parties, we know that the CDU is more conservative than the FDP, so it should be placed somewhere between the fixed points represented by the FDP and the AfD. Additionally, we recognize that the Grüne and SPD parties are more left-leaning than the FDP, indicating that they should be positioned between the fixed points of Die Linke and the FDP.

To begin our analysis with the party Die Grünen, we need to determine whether they align more closely with Die Linke or the FDP. To do so, we use the Wahlomat dataset described above. As it contains responses from political parties to the statements, we can compute the overlap across parties.

Consider the following scoring system for measuring agreement: assign a distance of 0.0 to two parties who provide identical answers, a distance of 0.5 to two parties whose responses differ slightly (one agrees or disagrees with a given statement while the other remains neutral), and a distance of 1.0 to two parties who are in complete disagreement. 
Figure \ref{fig:walomat-demonstration}a illustrates how the principle operates using the example of a particular election. Whenever there is an overlap in opinions, such as both parties endorsing the same statement, a striped pattern appears. 
The greater the number of striped boxes, the more similar the two parties are. In Figure \ref{fig:walomat-demonstration}b, we see that the Grüne party has the most overlaps with Die Linke. Meanwhile, Figure \ref{fig:walomat-demonstration}c accurately positions the Grüne party between Die Linke and the FDP, reflecting the calculated distances and angles.

The following calculation will be used for the sake of illustration. The Green Party and the Left Party collectively addressed 2,111 questions, providing identical responses to $I=1,530$ of them. In $P=284$ cases, one party took a neutral stance while the other either agreed or disagreed. Furthermore, on $O=297$ questions, the two parties expressed differing opinions.

The Green Party and the Liberal Party (FDP) answered a total of 2,249 questions together. They fully agreed on $I=828$ questions, partially agreed on $P=383$ questions, and disagreed on $O=1,038$ questions.

Let $d_{(a,b)}:=(0.5\cdot P+O)/T$ denote the estimated distance of two parties $a$ and $b$, where $T=I+P+O$. Regarding the example, this yields $d_{B'90,Linke}=0.208$ and $d_{B'90,FDP}=0.547$. The relative proximity of a party \textit{a} to the two reference parties \textit{b} and \textit{c} is then mapped onto the interval $[-90^\circ,0^\circ]$ for left wing resp. $[0^\circ,90^\circ]$ for right-wing parties by using $\theta_a = \varphi \cdot \big(d_{(a,b)}\big)/\big(d_{(a,c)}+d_{(a,b)}\big)$. Regarding the example of \textit{B'90 - die Grünen} being a potentially left party, $\varphi=-90^\circ$ is used, leading to $\theta_{B'90}\approx-65.2^\circ$.
The same reference parties are used for SPD, yielding $\theta_{SPD}\approx-53.9^\circ$. However, using $\varphi=90^\circ$ the more right-wing CDU party is compared with FDP and AfD, resulting in $\theta_{CDU}\approx37.9^\circ$. For implementation, the arctan2 function is used to determine the quadrant of the circle. See Table~\ref{tab:partyvectors} for an overview of all six party vectors.

Based on these angles $\theta_i$, we calculate unit vectors \textit{\textbf{v}\textsubscript{i}} for each party 
$i\in\{\mathrm{ Linke, B'90, SPD, FDP, CDU, AfD} \}$ as $\mathbf{v}_i := \big(\sin\left(\theta_{i}\right), \cos\left(\theta_{i}\right)\big)$.

\begin{table}[t]
\centering
%\resizebox{.95\columnwidth}{!}{
\begin{tabular}{lrrrrrr}
    \hline
        & Linke & B'90 & SPD & FDP & CDU & AfD  \\\hline
    $\theta^\circ$ & $-90.0$ & $-65.2$ & $-53.9$ & $0.0$ & $37.9$ & $90.0$ \\
    % $p$      & .031 & .281 &  .274 & .451 & .070 & .001 \\
    $v_x$    & $-1.0$ & $-0.9$ & $-0.8$ & $0.0$ & $0.6$ & $1.0$ \\
    $v_y$    &  $0.0$ &  $0.4$ & $0.6$ & $1.0$ & $0.8$ & $0.0$ \\\hline
\end{tabular}
\caption{Party vectors $\theta_i$ and unit vectors \textbf{\textit{v}}.}
\label{tab:partyvectors}
\end{table}

In the final step, the output $p_i$ from the multilabel classifier is multiplied by the corresponding vectors $\mathbf{v}_{i}$, and all of these vectors are summed together. Formally, $\mathbf{v}_{\mathrm{res}} = \sum_i p_i\,\mathbf{v}_{i}$. Finally, the angle is then calculated using $\mathrm{atan2}(\mathbf{v}_{\mathrm{res}})$ and divided through $\pi/2$ to transfer the resulting angle to a final classification score.

\subsection{Overall Architecture}
In this section, we combine all the building blocks that have been introduced in the previous sections. This is illustrated by processing the sentence: ``Familienpolitik soll Wahlfreiheit ermöglichen: gute Kitas, Ganztagsschulen und flexible Arbeitsmodelle.'' (Family policy should enable freedom of choice: good daycare centers, all-day schools, and flexible working models).
Since we are discussing social benefits, we expect a left-leaning result $(score < 0)$.

\subsubsection{Political Classifier} First, we check whether the example is political by using the DeBERTa political classifier introduced previously. This results in a score of 0.99, indicating that this statement is political.

\subsubsection{Political Party Classifier} If a text is classified as political, it is next processed using all 13 trained political party classifiers.
% Now the next step is put into one of the 13 trained political party classifiers. 
For this example, we use gemma2-9b solely, which yields the following probabilities: $P(\mathrm{party}=\mathrm{`Linke\textrm'})=.0307$, $P(\mathrm{`B'90\textrm'})=.2806$,  $P(\mathrm{`SPD\textrm'})=.2743$, $P(\mathrm{`FDP\textrm'})=.4508$, $P(\mathrm{`CDU\textrm'})=.0698$, $P(\mathrm{`AfD\textrm'})=.0011$.

\subsubsection{From Multilabel to a Continuous Scale}
We multiply the obtained model probabilities by the given party vectors and calculate the combined vector $\mathbf{v}_{\mathrm{res}} = (-0.159,  0.277)$.

From the combined result vector, we can accurately calculate the angle using the arctangent or atan2 function, which considers all four quadrants of the circle.

\[
\theta_{\mathrm{result}}
= \mathrm{atan2}\!\left(-0.159,\;0.277\right) \approx -0.521\ \mathrm{rad} \approx  -29.851^{\circ}
\]

To obtain a score $s$, scaled between -1 and 1 instead of -90 and 90, we must divide the result by 90 degrees or by $\pi/2$ when using radians: $\mathrm{score} = \theta_{result}^{\circ}/{90^{\circ}} \approx -0.332$.

Our example regarding family politics has resulted in a slightly left-leaning vector, as assumed previously.

\subsection{Evaluation Using Newspapers}
The final step of the evaluation involves using the newspaper ratings from Mediencompass.org to compare its general orientation with the classifier's results. For instance, the newspaper Bild \cite{springer_bild_2025} is rated 5.2 (0.4 on our scale, $[-1, 1]$) by the project, indicating a slightly right-wing orientation. Since our classifier operates at the per-article level, whereas the ground truth is at the per-newspaper level, we need to aggregate our classifier's results across all political articles and evaluate how closely the computed average aligns with the ground truth. As explained above, a filtering process is essential before assessing the political direction, ensuring that all articles meet a politicalness threshold of 0.8. If the average classification of all political articles in the newspaper Bild were 0.3, the error would be 0.1. The mean error across all 33 newspapers then reflects the quality of the specific trained classification model.

Let $a$ be an article of a newspaper $A$ out of the set of newspapers $\mathcal{A}$.

First, we compute the political leaning for each article $a$ in a newspaper $A$ and compare the result with the expected leaning based on Mediencompass.org $L$ (newspaper level). The mean absolute error $MAE$ of the tested model is computed as the average over the absolute differences between all newspapers tested by the respective model.

\subsection{Final Optimization}
\begin{figure}[!t]
% \begin{figure*}[!t]
  \centering
  \includegraphics[width=\linewidth]{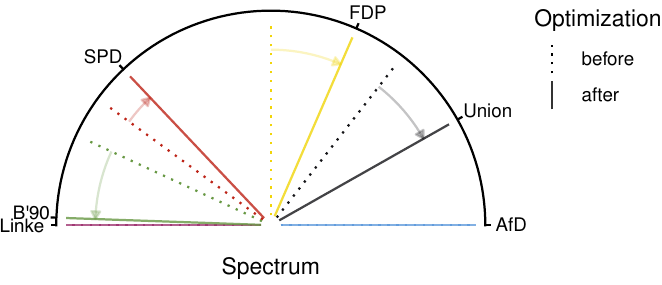}
  \caption{Comparison of the party vectors before and after the optimization for Gemma2-2b}
  \label{fig:optimization}
% \end{figure*}
\end{figure}

In our final step, we aim to refine the vector model to better align with the newspaper data. Specifically, while we optimize for the evaluation data, we impose a constraint that limits the adjustment of each party vector to a maximum of 0.25 in either direction. This new optimization builds on our initial use of the Wahlomat responses, in which we anchored the liberal party, the FDP, at the top of the semicircle. Although this initial method involved several guiding assumptions, it did not guarantee an 
%ensure an absolutely 
optimal outcome. By introducing these constraints, we acknowledge the value of our initial approach and seek to prevent the model from overfitting to the evaluation data.

Moreover, we position Die Linke on the left side and the AfD on the right side, while allowing for adjustments to the liberal FDP. This decision also has a technical basis. When we move the vectors representing the leftmost and rightmost positions upward, %there is no way to achieve these positions through a vector combination without moving backward
these positions cannot be reached through vector combinations without introducing negative contributions, which is not feasible, as the multilabel classifier only produces positive outputs ranging from 0 to 1. Thus, adjusting the leftmost and rightmost vectors would restrict the set of reachable vectors.

\[
\min_{\{\Delta v_p\}_{p \in \mathcal{P}}} \ \mathrm{MAE}(\tau) 
= \frac{1}{|\mathcal{A}|} \sum_{A \in \mathcal{A}} 
\left| \, \widehat L_A(\tau; \{v_p + \Delta v_p\}) - L_A \, \right|
\]

subject to $\|\Delta v_p\| \leq \delta_p \quad \forall\, p \in \mathcal P$
%\[
%\|\Delta v_p\| \le \delta_p \quad \forall\, p \in \mathcal P,
%\]

$$
\mathrm{with}~~ \delta_p = \left\{
\begin{array}{ll}
0    & \mbox{if } p \in \{\mbox{Linke},\, \mbox{AfD}\}\\[4pt]
0.25 & \mbox{else}
% \\
\end{array}
\right.
$$

\section{Results} % ca. 1.5 Seiten (davon 0.5 Plots)

% \begin{table}%[H]
% \centering
% \footnotesize
% \begin{tabular}{lrccc}
%   \hline
%   &&& \multicolumn{2}{c}{MSE/MAE}\\
%   \cline{4-5}
%   Classifier & Size & $F_1$ & pre & post \\ 
%   \hline
%   Gemma-2       &   2B & .75 & 0.053 / 0.185 & 0.043 / 0.172 \\
%   Gemma-2       &   9B & .82 & 0.047 / 0.186 & 0.043 / 0.172 \\
%   DeBERTa-l     & 435M & .84 & 0.055 / 0.197 & 0.047 / 0.182 \\
%   Llama-3.2     &   1B & .69 & 0.061 / 0.209 & 0.053 / 0.183 \\
%   GottBERT\_l   & 357M & .74 & 0.071 / 0.225 & 0.061 / 0.208 \\
%   gbert-large   & 337M & .73 & 0.062 / 0.196 & 0.062 / 0.192 \\
%   gelectra-l    & 336M & .72 & 0.066 / 0.211 & 0.067 / 0.202 \\
%   xlm-roberta-l & 561M & .71 & 0.080 / 0.225 & 0.080 / 0.218 \\
%   Llama-3.2     &   3B & .71 & 0.087 / 0.250 & 0.082 / 0.248 \\
%   gemma-3       &   1B & .69 & 0.202 / 0.371 & 0.125 / 0.287 \\
%   EuroBERT      & 610M & .79 & 0.249 / 0.419 & 0.133 / 0.304 \\
%   EuroBERT      & 210M & .73 & 0.143 / 0.304 & 0.133 / 0.299 \\
%   EuroBERT      & 2.1B & .72 & 0.146 / 0.305 & 0.144 / 0.307 \\
%    \hline
% \end{tabular}
% \caption{Overview of the used models, parameter sizes, evaluation metric ($F_1$), mean squared error (MSE), mean absolute error (MAE) before (pre) and after (post) optimization}
% \label{tab:results_final_old}
% \end{table}

\begin{table}%[H]
\centering
\footnotesize
\begin{tabular}{lrccccc}
  \hline
  && \multicolumn{2}{c}{pre}&& \multicolumn{2}{c}{post}\\
  \cline{3-4}\cline{6-7}
  Classifier & Size & MSE & MAE && MSE & MAE \\  %pre & post \\ 
  \hline
  Gemma-2       &   2B & 0.053 & 0.185 && 0.043 & 0.172 \\
  Gemma-2       &   9B & 0.047 & 0.186 && 0.043 & 0.172 \\
  DeBERTa-l     & 435M & 0.055 & 0.197 && 0.047 & 0.182 \\
  Llama-3.2     &   1B & 0.061 & 0.209 && 0.053 & 0.183 \\
  GottBERT\_l   & 357M & 0.071 & 0.225 && 0.061 & 0.208 \\
  gbert-large   & 337M & 0.062 & 0.196 && 0.062 & 0.192 \\
  gelectra-l    & 336M & 0.066 & 0.211 && 0.067 & 0.202 \\
  xlm-roberta-l & 561M & 0.080 & 0.225 && 0.080 & 0.218 \\
  Llama-3.2     &   3B & 0.087 & 0.250 && 0.082 & 0.248 \\
  gemma-3       &   1B & 0.202 & 0.371 && 0.125 & 0.287 \\
  EuroBERT      & 610M & 0.249 & 0.419 && 0.133 & 0.304 \\
  EuroBERT      & 210M & 0.143 & 0.304 && 0.133 & 0.299 \\
  EuroBERT      & 2.1B & 0.146 & 0.305 && 0.144 & 0.307 \\
   \hline
\end{tabular}
\caption{Pre versus post optimization comparison of the used models by mean squared error (MSE) and mean absolute error (MAE), ordered by post MSE}
\label{tab:results_final}
\end{table}

\subsection{Findings}
Upon reviewing the results, we conclude that our transformer models, when utilized with vectors, effectively identify political stances in German texts. The accuracy of our classifier closely aligns with that of public left/right polls, suggesting that its outcomes are consistent with those of human raters. Additionally, we found that a model’s size does not necessarily guarantee effectiveness across all scenarios, as depicted and further explained in Figure~\ref{fig:results-errorbars}. For instance, both Llama 3.2 models (with 1B and 3B parameters) performed worse on in- and out-of-domain classification than the significantly smaller DeBERTa-large model (435M). The DeBERTa-large model achieved the highest in-domain performance, with an $F_1$ score of 0.84, while Gemma2-2b demonstrated the best out-of-domain results both before and after optimization, as illustrated Table \ref{tab:results_final}.

\subsection{In-Domain Model Performance}
To evaluate the model's in-domain performance, we used the 20\% Bundestag and Wahlomat test set parts. The models that performed best were DeBERTa-large ($F_1=0.84$), Gemma-2-9b ($F_1=0.79$), and EuroBERT-610m ($F_1=0.79$).

\subsection{Out-of-Domain Model Performance on Tweets}
The out-of-domain evaluation was carried out using posts from members of the German Bundestag. It is important to note that our knowledge of each tweet is limited to its author and their associated party; we do not have information about the tweet's political stance or how it might be received by others. Consequently, our evaluation focused solely on the accuracy of the three top-performing classifiers from the earlier in-domain classification task.
We investigated a strong correlation between tweet length and the accuracy of our classifier of $0.96\leq r \leq 0.97$ depending on the used model. This aspect is crucial, as tweets are often quite brief, and in some cases, the author may require the reader to rely on external context to fully understand a quote.

\begin{figure}[!t]
  \centering
  \includegraphics[width=1\linewidth]{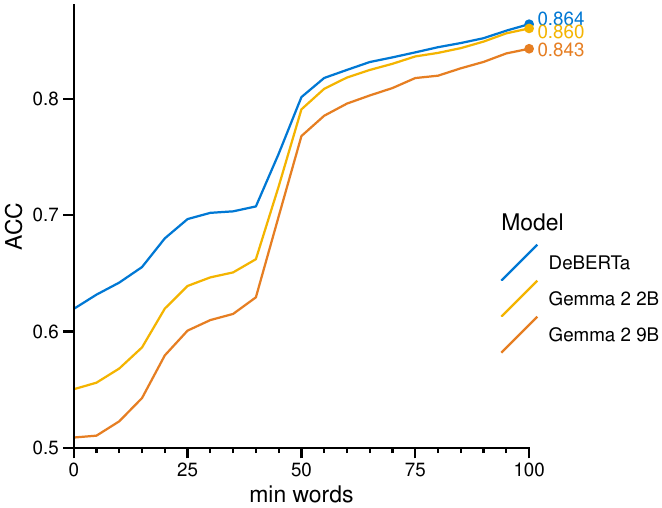}
  \caption{Classifier performance on tweets}
  \label{fig:tweet_performance}
\end{figure}

As illustrated in Figure \ref{fig:tweet_performance}, the accuracy for shorter tweets ranges between 50\% and 65\%. However, this accuracy increases to over 80\% when tweets contain 50 or more words.

\subsection{Out-of-Domain Model Performance on Newsmedia}
The out-of-domain evaluation on newspapers was conducted using 33 German news outlets, ranging from left-wing publications like Jungle World to right-wing ones, such as Compact. Additionally, only news articles with a political score of 0.8 or higher were further processed and reviewed using a different model. The score was calculated as the MAE combined with the percentage relative to the total scale of 2 [-1, 1], providing a clearer perspective. The top-performing models were Gemma2-2b with an MAE of 0.1852 (9.26\%), Gemma2-9b with an MAE of 0.1859 (9.29\%), and gbert-large with an MAE of 0.1965 (9.82\%). Notably, the ranking differs from the in-domain results, suggesting that some models generalize text more effectively than others.

\subsection{Effect of the Vector Optimization}
Following the initial out-of-domain test, we aimed to determine whether a different vector alignment could minimize the mean absolute error. 
%mean average error. 
We conducted a numerical optimization that allowed us to adjust each vector by ±0.25, except for the extreme positions held by Die Linke and AfD, which had to remain fixed to ensure the model's functionality. Our findings revealed that the optimizer successfully minimized the mean absolute error while keeping the adjustments within the specified range. Figure \ref{fig:optimization} illustrates the party vectors before and after the optimization process. Notably, the Grüne party shifted further to the left, while the FDP and CDU moved to the right. Post-optimization, each model possessed its own tailored set of vectors for all six parties.

After optimization, all models exhibited an average decrease in mean absolute error of 0.0239, corresponding to 1.1946\%. EuroBERT-610m demonstrated the most substantial reduction, with a decrease of 0.304 or 5.73\%, whereas EuroBERT-2.1B was the only model to perform worse, with a decline of 0.0016 or 0.08\%.

\section{Discussion} % ca. 1.5 Seiten 

To train multiple BERT, Llama, and Gemma LLMs, a wide variety of data was used, including minutes from the German Bundestag and statements from a digital voters' guide (Wahlomat). Unlike the Bundestag data, which captures attitudes implicitly (e.g., via emotions), the Wahlomat dataset is based on explicit information provided directly by the parties. In order to further increase the (linguistic) variance, both datasets were artificially enriched. The models' multi-label responses were translated into a numerical continuum from -1 (left) to 1 (right). Performance was tested in (test set) and out of the domain (independent newspaper dataset and tweets).

We have observed that a trained multi-label transformer model, when combined with the appropriate party vector projection, can recognize political stances at a level comparable to that of polls for political classification. During our search for the optimal transformer model for our classifier, we discovered that the best-performing model within the specific domain (DeBERTa-large with an F1 score of 0.84) does not necessarily translate to being the best for other domains. In the out-of-domain test, DeBERTa-large was outperformed by Gemma-2-2B, indicating that a combination of model size, architecture, and training data is crucial for achieving superior out-of-sample accuracy. 

The German-pretrained DeBERTa-large model demonstrated superior performance in in-domain classification, attributable to its extensive training corpus, as detailed in the models section of the methodology chapter. Unlike the conventional approach of relying solely on the OSCAR dataset \cite{ortiz_suarez_asynchronous_2019}, the developers of the German DeBERTa model \cite{dada_impact_2023} leveraged a more diverse dataset encompassing multiple domains. This strategic selection, incorporating formal, informal, legal, medical, and literary texts, enhanced the model's in-domain efficacy, underscoring the importance of data diversity in model training.

A compelling reason for the outstanding performance of the Gemma2-2b and Gemma2-9b models \cite{riviere_gemma_2024} in the out-of-domain classification task is their unique training paradigms. By employing knowledge distillation from larger models, these models function as effectively condensed experts. This approach enables them to generalize significantly better than other models within the same size category.

Additionally, we found that providing the vectors from our initial approach to an optimizer, allowing it some flexibility in adjusting those vectors, can further enhance out-of-domain accuracy. Our method represents a novel approach compared to many existing techniques, which rely on discrete labels.
Despite the evolving political landscape, our proposed methodology eliminates the need for manual labeling. This enables us to conduct periodic retraining by updating only the training datasets for our models.

\subsection{Practical Implications}
It is important to acknowledge that individuals exhibit biases to varying degrees. %everyone has their biases to some degree, and that this is completely normal. 
However, being aware of these biases can significantly enhance our understanding of the world. A classifier like the one introduced in this paper can assist in categorizing news outlets, authors, discussion threads, and various conversations, helping to prevent us from becoming trapped in echo chambers and encouraging a broader perspective. Additionally, a browser plugin could display the bias of every newspaper we visit.

An implementation could involve tracking a rolling score over a day or week to monitor the trajectory of topics, newspapers, or discussion threads. Warning signals could be triggered if a news outlet remains too entrenched in one extreme for an extended period.

Another application could involve conducting targeted searches for discussions aligned with left- or right-wing ideologies to analyze the topics with which specific groups engage. This would facilitate a deeper understanding of discourse patterns and sentiment within these ideological communities. 

Our approach is highly adaptable to different countries and use cases because we avoid manual labeling and extract the political spectrum directly from the data. For instance, if there is a political shift in Germany, we can easily accommodate this by retraining the model. In contrast, comparable projects that rely on manual labeling require a significantly larger workforce. Our primary requirements are political texts from parties and fixed reference points, such as newspapers. A well-distributed representation of political parties within the spectrum is particularly beneficial, as it facilitates coverage of various points by combining those vectors. In contexts with only two major parties, achieving fine-grained classification can be more challenging because there is no liberal segment between them.

Our paper can also serve as a foundation for further research on social media. For instance, if a researcher has collected sufficient social media data and aims to track political shifts before, during, and after a governmental transition, they can use our tool. 

Adapting the tool for other languages and cultural contexts is also possible. 
We introduce a new method for effectively classifying texts along the political spectrum, using comments and other indicators as proxies to construct a training corpus. 

\subsection{Limitations}

The limitations can be categorized as model-related and methodological limitations. In our case, model-related limitations arise from using a classification model rather than a reasoning-based transformer model. At times, when quoting individuals, it can be challenging for the model to discern whether the quoted opinion is being critiqued. For example, consider the tweet: `` \lq Those who want human society must overcome male society.\rq Svenja Schulze (February 17, 2022) This was or is also stated in the SPD party program.''. Based on the available information, we cannot determine whether the tweet's author agrees or disagrees with the statement. The classifier will categorize it as left-wing solely on the basis of the quoted content.

Another limitation of our analysis's non-reasoning capabilities lies in its evaluation of foundational statements. For instance, when we input Pierre-Joseph Proudhon's statement ``Property is theft'' \cite{proudhon_quest-ce_1840} into our gemma-2-9b classifier, it produces a score of 0.78, classifying it as a right-wing assertion. Despite Proudhon's typically being recognized as a figure of the libertarian left \cite{honeywell_anarchism_2021, levy_palgrave_2019}, some scholars have controversially interpreted him as a precursor to fascism \cite{schapiro_pierre_1945, krier_sozialismus_2009}. In this scenario, it would be intriguing to understand why the classification model categorized the statement as right-wing. However, because we are not using a reasoning model, we cannot examine the rationale for the classification. 
A possible explanation for the misclassification in this example is that such ideological statements are neither discussed in parliament nor part of the political agenda of mainstream political parties.

Given the increase in accuracy associated with a higher number of words per tweet, we conclude that the models may struggle with very short texts, particularly when background knowledge is required to interpret their meaning.

From a methodological standpoint, a one-dimensional projection may not offer enough entropy to accurately map political views in certain instances. This becomes evident when we examine Figure \ref{fig:sentiment-correlations}. One might intuitively assume that the left-most party, Die Linke, and the right-most party, AfD, exhibit the greatest distance. However, the negative correlations between the right-wing AfD and the Grüne (-0.32) and SPD (-0.34) parties are significantly stronger than that between the AfD and Die Linke (-0.18).

One possible interpretation for this phenomenon is that both the AfD and Die Linke respond to certain issues in similar ways, albeit for distinct reasons. For instance, consider the issue of supplying arms to a nation under attack by its neighbor. A left-wing party such as Die Linke would oppose these arms deliveries from a pacifist perspective, whereas the right-wing AfD would object from a nationalist perspective, prioritizing Germany’s interests and its trade relations with the aggressor state.

Another possible explanation is that both the left-wing Die Linke and the right-wing AfD are not part of the governing coalitions and instead belong to the opposition. In this role, opposition parties commonly critique the governing parties' policies and actions. Consequently, this can lead to the formation of similar opinions on both ends of the political spectrum.

The political shift occurring in numerous societies has been noted, but it also presents limitations. What may be viewed as slightly left or right today could be regarded as a liberal stance in the near future. To address this issue, it is recommended that the classifier be retrained with updated data whenever such a shift becomes apparent or at regular intervals.

A classifier's applicability is inherently limited to the linguistic and cultural context in which it was trained. For instance, the fundamentally divergent perspectives on public health care and labor rights observed in Germany and the United States underscore the pitfalls of deploying a classifier across different cultural frameworks without appropriate adaptation.
The cultural context not only establishes limits for the classifier, but the diversity of the input data also plays a significant role. Positions further to the left of the party Die Linke, as well as those to the right of the party AfD, cannot be distinguished from the positions of these two parties since the scale based on the six parties is bounded by these parties.

It is essential to note that we employ a classification-based approach, which inherently lacks interpretability and reasoning. The classifier cannot provide insights into why it categorizes a specific text as belonging to the Grüne or SPD parties. Additionally, it does not explicitly model underlying principles of left- or right-wing politics, as it is not designed to serve as a reasoning model. While it effectively categorizes topics in the training data, it may struggle to classify novel concepts, as it lacks the capacity to reason about them.
In summary, while the classifier can make errors, it should not be relied upon to block texts without additional verification or similar measures. To minimize the impact of misclassification, it is advisable to apply the classifier to a large volume of texts, such as those in a newspaper, as illustrated in the given example. This way, the significance of any single mistake is diminished.

\subsection{Future Work}
To address the limitations discussed above, it would be advantageous to develop a classifier capable of explaining why it assigns a text to a specific position on the left-right spectrum. This feature would help users understand the rationale for classifying a text as far-right or far-left and would also facilitate the identification of potential errors. Users could contest the classifier's assessment and dismiss the result in cases of inaccuracies, rather than accepting its conclusions unquestioningly. Additionally, a reasoning model would enhance its utility by enabling it to apply foundational concepts of left, liberal, and right-wing politics to new topics not encountered in its training data. 
Improved explainability and generalizability could benefit future developments.

\subsection{Social Impact and Misuse}
%Did you discuss any potential negative societal impacts of your work
%Did you discuss any potential misuse of your work?
%id you describe steps taken to prevent or mitigate potential negative outcomes of the research, such as data and model documentation, data anonymization, responsible release, access control, and the reproducibility of findings?
The developed models, once sufficiently advanced for everyday use, could potentially pose risks if individuals rely on them to filter news. Instead of expanding their perspectives, users might choose to exclude all news that exceeds a certain threshold in a political direction, whether left or right. As a result, what is intended to be a useful application could also be misused.

A second potential misuse of the model is the risk of discrimination against individuals based on their political beliefs. On a broader level, institutions could utilize the model to monitor live social media streams and identify individuals expressing dissenting opinions.

We firmly oppose all forms of discrimination on the basis of political viewpoints and strongly advocate for free speech. Additionally, we are committed to transparency by making all training data publicly available, ensuring that our models and their results are clear and accessible.

\section*{Ethical Statement}
All of our training data has been sourced exclusively from publicly available materials and is intended solely for academic use. We did not bypass any safety mechanisms or use data behind a paywall. Additionally, we have not collected any personal data, except for political speeches by public figures.

We emphasize that the introduced classification method is still in its early stages, and errors cannot be discounted. Furthermore, our classifier should not be used to evaluate or discriminate against individuals on the basis of their political beliefs. We strongly advocate for a diverse political landscape and uphold the principles of free speech in respectful interactions, free from personal attacks.

\section*{Acknowledgments}
The authors would like to thank the System Sciences Chair for Communication Systems and Network Security under the direction of Prof. Dr. Gabi Dreo Rodosek.

The authors acknowledge the financial support from the Federal Ministry of Education and Research of Germany in the program “Souverän. Digital. Vernetzt.” Joint project 6G-life, project identification number: 16KISK002.

\bibliography{aaai2026}

\section{Paper Checklist}

\begin{enumerate}

\item For most authors...
\begin{enumerate}
    \item  Would answering this research question advance science without violating social contracts, such as violating privacy norms, perpetuating unfair profiling, exacerbating the socio-economic divide, or implying disrespect to societies or cultures?
    \answerYes{Yes, see the Ethical Statement}
  \item Do your main claims in the abstract and introduction accurately reflect the paper's contributions and scope?
    \answerYes{Yes}
    % {Yes, we even included problem, approach, and contribution in the introduction}
   \item Do you clarify how the proposed methodological approach is appropriate for the claims made? 
    \answerYes{Yes, see the Methods section}
    % {Yes, in the methodology section and more precisely in section ``From multilabel to a continuous scale''}
   \item Do you clarify what are possible artifacts in the data used, given population-specific distributions?
   \answerYes{Yes, see the Methods, Subsection \textit{Dataset}}
    % {Yes, we talked about our data selection in the dataset section}
\item Did you describe the limitations of your work?
  \answerYes{Yes, see the Discussion, Subsection \textit{Limitations}}
    % {Yes, spend half a page of our results section talking about limitations}
\item Did you discuss any potential negative societal impacts of your work?
  \answerYes{Yes, see the Discussion, Subsection \textit{Practical implications}}
    % {Yes, please see section social impact and misuse}
\item Did you discuss any potential misuse of your work?
    \answerYes{Yes, see the Discussion, Subsection \textit{Social impact and misuse}}
    %{Yes, please see section social impact and misuse}
\item Did you describe steps taken to prevent or mitigate potential negative outcomes of the research, such as data and model documentation, data anonymization, responsible release, access control, and the reproducibility of findings?
    \answerYes{Yes, see the Discussion, Subsection \textit{Social impact and misuse}}
    % {Yes, please see section social impact and misuse}
\item Have you read the ethics review guidelines and ensured that your paper conforms to them?
    \answerYes{Yes}
\end{enumerate}

\item Additionally, if your study involves hypotheses testing...
\begin{enumerate}
  \item Did you clearly state the assumptions underlying all theoretical results?
    \answerNA{NA}
  \item Have you provided justifications for all theoretical results?
    \answerNA{NA}
  \item Did you discuss competing hypotheses or theories that might challenge or complement your theoretical results?
    \answerNA{NA}
  \item Have you considered alternative mechanisms or explanations that might account for the same outcomes observed in your study?
    \answerNA{NA}
  \item Did you address potential biases or limitations in your theoretical framework?
    \answerNA{NA}
  \item Have you related your theoretical results to the existing literature in social science?
    \answerNA{NA}
  \item Did you discuss the implications of your theoretical results for policy, practice, or further research in the social science domain?
    \answerNA{NA}
\end{enumerate}

\item Additionally, if you are including theoretical proofs...
\begin{enumerate}
  \item Did you state the full set of assumptions of all theoretical results?
    \answerNA{NA}
	\item Did you include complete proofs of all theoretical results?
    \answerNA{NA}
\end{enumerate}

\item Additionally, if you ran machine learning experiments...
\begin{enumerate}
\item Did you include the code, data, and instructions needed to reproduce the main experimental results (either in the supplemental material or as a URL)?
  \answerYes{Yes}
    % {Yes, the reference will be temporarily commented out to avoid jeopardizing the double-blind publication process.}
\item Did you specify all the training details (e.g., data splits, hyperparameters, how they were chosen)?
  \answerYes{Yes, see the linked GitHub repository, folder 05\_train\_new\_model}
    % {Yes, they are all part of the training files in the GitHub repository, folder: 05\_train\_new\_model}
\item Did you report error bars (e.g., with respect to the random seed after running experiments multiple times)?
    \answerYes{Yes, see Figure~\ref{fig:results-errorbars}}
    % \answerYes{Yes, the models were trained several times, see Figure \ref{fig:f1_eval}}
    % {Yes we trained the models several times, see Figure 3: Multiple runs of the DeBERTa-large model}
\item Did you include the total amount of compute and the type of resources used (e.g., type of GPUs, internal cluster, or cloud provider)?
    \answerYes{Yes, see Table \ref{tab:results}}
    % {Yes, see Table 2: Overview of the used models, parameter sizes, evaluation metrics, training hours, and used GPUs.}
\item Do you justify how the proposed evaluation is sufficient and appropriate to the claims made?
    \answerYes{Yes, see the Methods}
    % {Yes, we ran small models multiple times and stopped larger ones when the loss stopped decreasing. Further, we also evaluate using out-of-sample data.}
\item Do you discuss what is ``the cost`` of misclassification and fault (in)tolerance?
\answerYes{Yes, see the Discussion, Subsection \textit{Limitations}}
    % {Yes, we discuss this in our limitations section, where we recommend applying the classifier to a high number of texts.}
  
\end{enumerate}

\item Additionally, if you are using existing assets (e.g., code, data, models) or curating/releasing new assets, \textbf{without compromising anonymity}...
\begin{enumerate}
\item If your work uses existing assets, did you cite the creators?
    \answerYes{Yes, see the References and the Methods, Subsection \textit{Dataset}}
    % {Yes, we utilized the plenary protocols and the Wahlomat dataset, both of which we cite as the original source in our dataset section.}
\item Did you mention the license of the assets?
    \answerYes{Yes}
    % {Yes, we wrote ``The plenary protocols of the German Bundestag (Deutscher Bundestag 2025) are considered open data''}
\item Did you include any new assets in the supplemental material or as a URL?
    \answerYes{Yes, we included links to the code and datasets}
    % {Yes, we included huggingface links to our own created datasets, but we will anonymize them for the peer review}
\item Did you discuss whether and how consent was obtained from people whose data you're using/curating?
  \answerYes{Yes, see the Ethical Statement}
    % {Yes, we state in our ethical statement that we are using speeches from politicians, but that these politicians are considered public figures.}
\item Did you discuss whether the data you are using/curating contains personally identifiable information or offensive content?
  \answerYes{Yes, see the Ethical Statement the Methods, Subsection Datasets}
    % {Yes, we mention in our newspaper dataset section that some newspapers might contain offensive content, but we intentionally leave them unchanged to allow the later classifiers to learn from the content. The plenary protocols contain personal identifiable information of the politicians, but they are public figures.}
\item If you are curating or releasing new datasets, did you discuss how you intend to make your datasets FAIR (see \citet{fair})?
    \answerYes{Yes}
% {Yes, we made the data available using HuggingFace and assigned a DOI to it, so it can't be taken down and the DOI Version will stay the same}
\item If you are curating or releasing new datasets, did you create a Datasheet for the Dataset (see \citet{gebru2021datasheets})? 
    \answerYes{Yes, see the model cards on HuggingFace}
% {Yes, the model card on HuggingFace for our curated dataset from the plenary protocols and Wahlomat data is based on the work of \citeauthor{gebru2021datasheets}.}
\end{enumerate}

\item Additionally, if you used crowdsourcing or conducted research with human subjects, \textbf{without compromising anonymity}...
\begin{enumerate}
  \item Did you include the full text of instructions given to participants and screenshots?
    \answerNA{NA}
  \item Did you describe any potential participant risks, with mentions of Institutional Review Board (IRB) approvals?
    \answerNA{NA}
  \item Did you include the estimated hourly wage paid to participants and the total amount spent on participant compensation?
    \answerNA{NA}
   \item Did you discuss how data is stored, shared, and deidentified?
   \answerNA{NA}
\end{enumerate}

\end{enumerate}

% \bibliography{aaai25}

% \end{document}

\appendix 
\section{Appendices}
\subsection{Tables and Figures}
% \subsection{A Tables}

\begin{table}[h]
  % \scriptsize
  \centering
  \begin{tabular}{ccccccc}
    No & Linke & B'90 & SPD & FDP & CDU & AfD \\
    \hline
    1 & yes & yes & yes & no & no & no \\
    2 & yes & yes & yes & yes& yes& no \\
    3 & yes & yes & yes & no & no & no \\\hline
  \end{tabular}
  \caption{(Dis)agreement of various parties regarding three exemplary statements: 1. \textit{A tax is to be reintroduced on high net worth individuals}, 2. \textit{Germany should keep the euro as its currency.}, 3. \textit{A minimum wage should be introduced.}}
  \label{tab:paraphrasing}
\end{table}

\begin{table*}
\centering
\begin{tabular}{lrrcrrrrcrrrr}
  \hline
  % \toprule % A = Kompas, B = Maurer, C = factcheck
    & \multicolumn{2}{c}{Source A} && \multicolumn{4}{c}{Source B} && \multicolumn{3}{c}{Source C}\\
  \cline{2-3}\cline{5-8}\cline{10-12}
  Media                  & $x$\;\; & $z(x)$\! && $x$\;\; & $y$\;\; & $PC_{1}$\! & $z(PC)$ && ord.\! & $x$\;\; & $z(x)$\! & Size\; \\ 
  \hline % \midrule
  Achgut                 & 5.20 &  1.19 &&      &      &      &        &&    &       &       & 72.1K \\ 
  Augsburger Allgemeine  &      &       && 2.74 & 2.87 &  0.46 &  0.46 &&    &       &       &  \\ 
  Bayerische Rundfunk    & 4.40 &  0.39 && 2.62 & 3.17 &  0.69 &  0.69 &&  0 & -1.70 & -0.62 & 40.7K \\ 
  Berliner Zeitung       &      &       && 2.47 & 2.66 & -0.17 & -0.17 && -1 & -3.00 & -1.01 &  \\ 
  Bild (Springer)        & 5.20 &  1.19 && 2.89 & 3.49 &  1.46 &  1.46 &&  1 &  3.60 &  0.97 & 1.47M \\ 
  Cicero                 & 4.90 &  0.89 &&      &      &       &       &&  1 &       &       & 13.5K \\ 
  Compact                & 6.00 &  1.99 &&      &      &       &       &&    &       &       & 11.8K \\ 
  der Freitag            & 2.70 & -1.31 &&      &      &       &       &&    &       &       & 26.8K \\ 
  Deutschlandfunk        & 3.80 & -0.21 && 2.94 & 2.44 &  0.18 &  0.19 &&  0 &       &       & 89.0K \\ 
  FAZ                    & 4.50 &  0.49 && 2.85 & 2.47 &  0.11 &  0.11 &&  1 &  3.40 &  0.91 & 316.0K \\ 
  Focus                  & 4.90 &  0.89 && 3.71 & 2.89 &  1.78 &  1.79 &&  1 &  2.80 &  0.73 & 33.5K \\ 
  Frankfurter Rundschau  & 3.40 & -0.61 && 2.22 & 1.98 & -1.38 & -1.38 && -1 &       &       & 165.0K \\ 
  General-Anzeiger Bonn  &      &       && 2.67 & 2.75 &  0.22 &  0.22 &&  0 &  1.50 &  0.34 &  \\ 
  Handelsblatt           & 4.30 &  0.29 && 3.69 & 2.58 &  1.36 &  1.36 &&  1 &  2.10 &  0.52 &  \\ 
  junge Freiheit         & 5.80 &  1.79 && 3.14 & 3.97 &  2.40 &  2.41 &&    &       &       & 56.0K \\ 
  junge Welt             & 2.40 & -1.61 && 1.84 & 2.14 & -1.67 & -1.67 && -2 &       &       & 91.7K \\ 
  Jungle World           & 2.30 & -1.71 &&      &      &       &       &&    &       &       & 57.7K \\ 
  Linksunten (indymedia) & 2.00 & -2.01 &&      &      &       &       &&    &       &       & 74.0K \\ 
  MDR                    & 4.10 &  0.09 && 2.69 & 2.35 & -0.28 & -0.28 &&    &       &       & 63.2K \\ 
  MM News                & 5.10 &  1.09 &&      &      &       &       &&    &       &       & 197.0K \\ 
  Münchner Merkur        &      &       && 2.75 & 2.98 &  0.61 &  0.61 &&  0 &  1.50 &  0.34 &  \\ 
  NDR                    & 3.70 & -0.31 &&      &      &       &       &&    &       &       & 35.2K \\ 
  Neues Deutschland      & 2.60 & -1.41 && 1.71 & 1.80 & -2.28 & -2.29 && -2 &       &       & 283.0K \\ 
  NTV                    & 4.30 &  0.29 && 2.56 & 2.60 & -0.13 & -0.13 &&  1 &       &       & 548.0K \\ 
  NachDenkSeiten         & 3.10 & -0.91 &&      &      &       &       &&    &       &       & 20.4K \\ 
  RT Deutsch             & 5.10 &  1.09 &&      &      &       &       &&    &       &       & 44.2K \\ 
  RTL                    & 4.50 &  0.49 && 2.50 & 3.14 &  0.49 &  0.49 &&    &       &       & 187.0K \\ 
  Rheinische Post        &      &       && 2.49 & 2.49 & -0.36 & -0.36 &&  1 &  2.30 &  0.58 &  \\ 
  Saarbrücker            &      &       && 2.72 & 2.76 &  0.30 &  0.30 &&  0 & -0.70 & -0.32 &  \\ 
  Spiegel                & 3.50 & -0.51 && 2.40 & 2.37 & -0.63 & -0.63 && -1 &       &       & 1.11M \\ 
  Stern                  & 3.80 & -0.21 &&      &      &       &       &&    &       &       & 414.0K \\ 
  Süddeutsche            & 3.50 & -0.51 && 2.36 & 2.34 & -0.73 & -0.73 && -1 & -3.40 & -1.13 & 1.89M \\ 
  T-Online               &      &       && 2.75 & 2.62 &  0.16 &  0.16 && -1 & -2.20 & -0.77 &  \\
  TAZ                    & 2.80 & -1.21 && 1.86 & 1.98 & -1.86 & -1.87 && -2 &       &       & 725.0K \\ 
  Tagesschau (ARD)       & 3.70 & -0.31 && 2.59 & 2.66 &  0.00 &  0.00 &&  0 &       &       & 55.8K \\ 
  Tagesspiegel           & 3.60 & -0.41 && 2.50 & 2.74 & -0.03 & -0.03 &&    &       &       & 1.20M \\ 
  Tichys                 & 5.50 &  1.49 && 3.62 & 3.99 &  3.07 &  3.08 &&  2 &  6.80 &  1.93 & 33.3K \\ 
  Vice                   & 2.80 & -1.21 &&      &      &       &       &&    &       &       & 53.8K \\ 
  WDR                    & 3.50 & -0.51 && 2.00 & 2.73 & -0.70 & -0.70 &&    &       &       & 45.8K \\ 
  Web.de                 &      &       && 2.56 & 2.60 & -0.12 & -0.12 && -1 & -2.20 & -0.77 &  \\
  Welt                   & 4.80 &  0.79 && 3.11 & 3.29 &  1.49 &  1.50 &&  1 &  3.60 &  0.97 & 164.0K \\ 
  Zeit                   & 3.60 & -0.41 && 2.47 & 2.71 & -0.11 & -0.11 && -1 &       &       & 343.0K \\   
  \hline % \bottomrule

\end{tabular}

\caption{Overview of the German media landscape, including several online versions of newspapers like \textit{Frankfurter Allgemeine Zeitung} (FAZ), \textit{die tageszeitung} (TAZ); television channels like \textit{Mitteldeutscher Rundfunk} (MDR), \textit{Norddeutscher Rundfunk} (NDR), \textit{Westdeutsche Rundfunk} (WDR), \textit{Radio Télévision Luxembourg} (RTL), and various other online news media formats. Media bias estimates were collected from three sources: 
A. ratings of $k=39$ media outlets from $n=1148$ participants on a seven-point Likert scale from \textit{extrem left} (1) to \textit{extrem right} (7), provided by \citet{Medienkompass_2025}; B. ratings of $k=47$ media outlets from only $n=9$ extensively trained raters on two correlated five-point scales, provided by \citet{maurer_fehlt_2024}; and C. ratings regarding $k=77$ outlets rated on a scale from \textit{extrem left} (-10) to \textit{extreme right} (10) retrieved from \citet{mediabiasfactcheck_2025}. For source B, both correlated scales were reduced using principal components analysis (PCA), yielding one principal component (PC). Numeric ratings were \textit{z}--transformed for comparability (standardised, i.e. $M=0$, $SD=1$). Regarding source C, numeric scores were missing for various media outlets; ordinal (ord.) scores were estimated from the given labels accordingly. Appropriate association estimates for pairwise completed cases showed high correlations, indicating convergent validity. Based on the media list from source A, $k=33$ (approx. 85\%) media outlets were scraped, yielding a dataset of approx. 10M articles; 74K per outlet on average. % tab:full_medienkompass
}
% Mediencompass.org, and mediabiasfactcheck.com 
% min	12K			
% mdn	74K			
% max	2M
% was measured on a seven point scale from \textit{left-wing extremist} (1) to \textit{right-wing extremist} (7); resp. quality on a five point scale from low as \textit{clickbait} (1) to  \textit{complex} (5). Data from \citeauthor{Medienkompass_2025} \shortcite{Medienkompass_2025}, based on a sample of $N=1148$ participants

\label{tab:full_medienkompass}
\end{table*}

\begin{table*}
  % \scriptsize
  \centering
  \begin{tabular}{lp{15cm}l}
    \hline
    Type & Statement  \\\hline
    original & A tax is to be reintroduced on high net worth individuals. \\
    child & Soon there will be a tax again for people who have a lot of money. This means that they will have to give part of their money to the government. The government can then use this money to buy important things for everyone.\\
    teenager & The federal government plans to reintroduce a tax on high net worth individuals. This means that people with large fortunes will have to hand over part of their wealth to the state. The revenue will be used for important public tasks. \\
    adult & The reintroduction of a wealth tax for high net worth individuals is currently under discussion. This measure is intended to help increase government revenue and reduce social inequality. By taxing large fortunes, the government can finance important investments in infrastructure and social programs. \\
    eloquent & There are growing calls for a tax to be levied on high net worth individuals in order to combat social inequality in our society. This measure could make an important contribution to the redistribution of wealth. \\
    tweet & A wealth tax would also stimulate the economy by promoting consumption and investment. A positive effect for everyone!\\\hline
  \end{tabular}
  \caption{Example of paraphrasing an original statement in the words of different persona}
  \label{tab:paraphrasing2}
\end{table*}

% \subsection{B Figures}

\begin{figure}
  \centering
  \includegraphics[width=.90\linewidth]{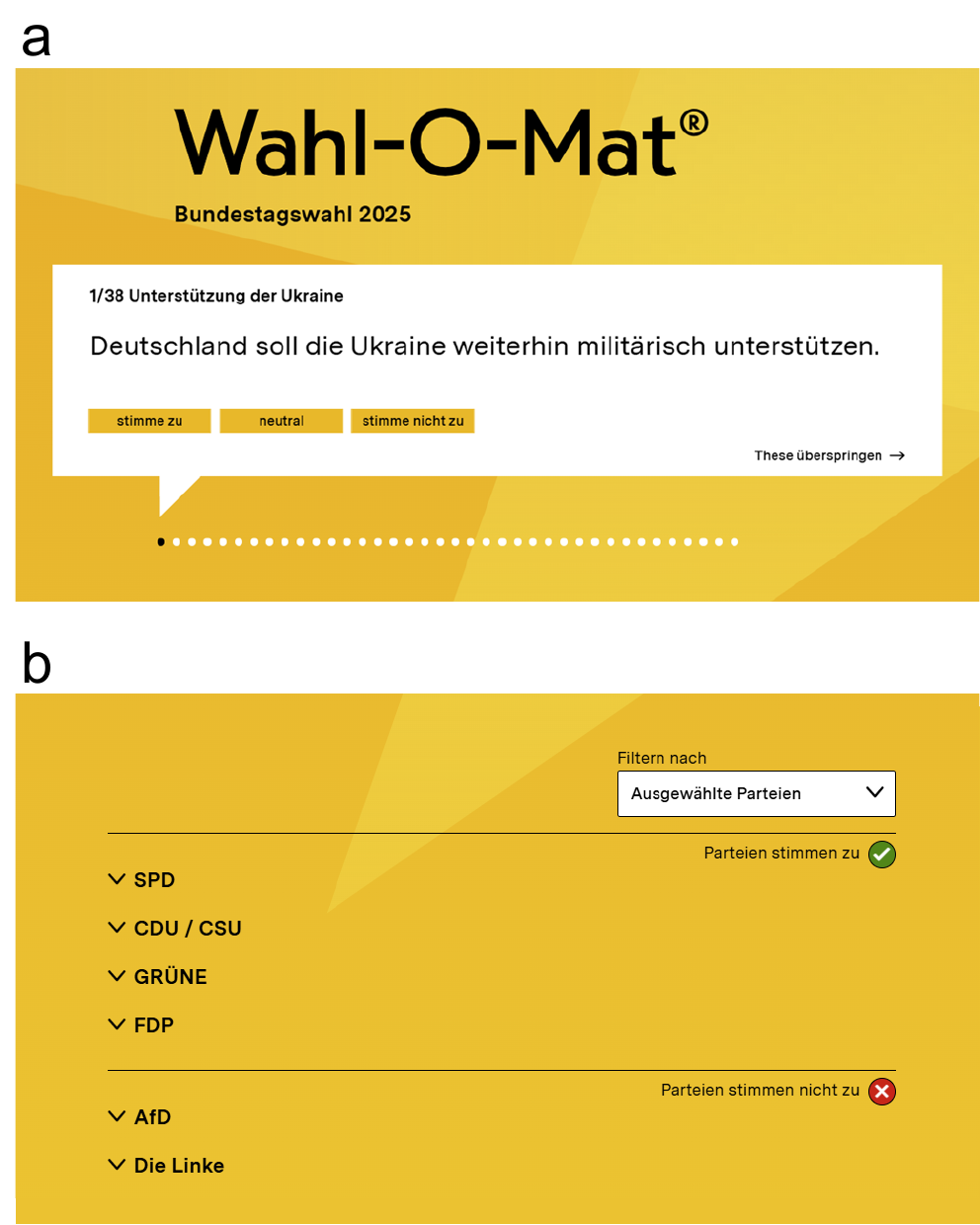}
  \caption{Exemplary statement 1/38: \textit{Germany should continue to provide military support to Ukraine}, sourced from the Wahlomat service regarding the German federal elections in 2025 (www.wahl-o-mat.de/bundestagswahl2025). Screenshot {\sffamily\small a} shows the user view with response options (approval, neutral, disapproval), {\sffamily\small b} depicts the stance of selected parties (disapproval by the most left-wing and right-wing parties, approval by the others).}
  \label{fig:wahlomat-comb}
\end{figure}

\begin{figure} %[htbp]
  \centering
  \includegraphics[width=.9\linewidth]{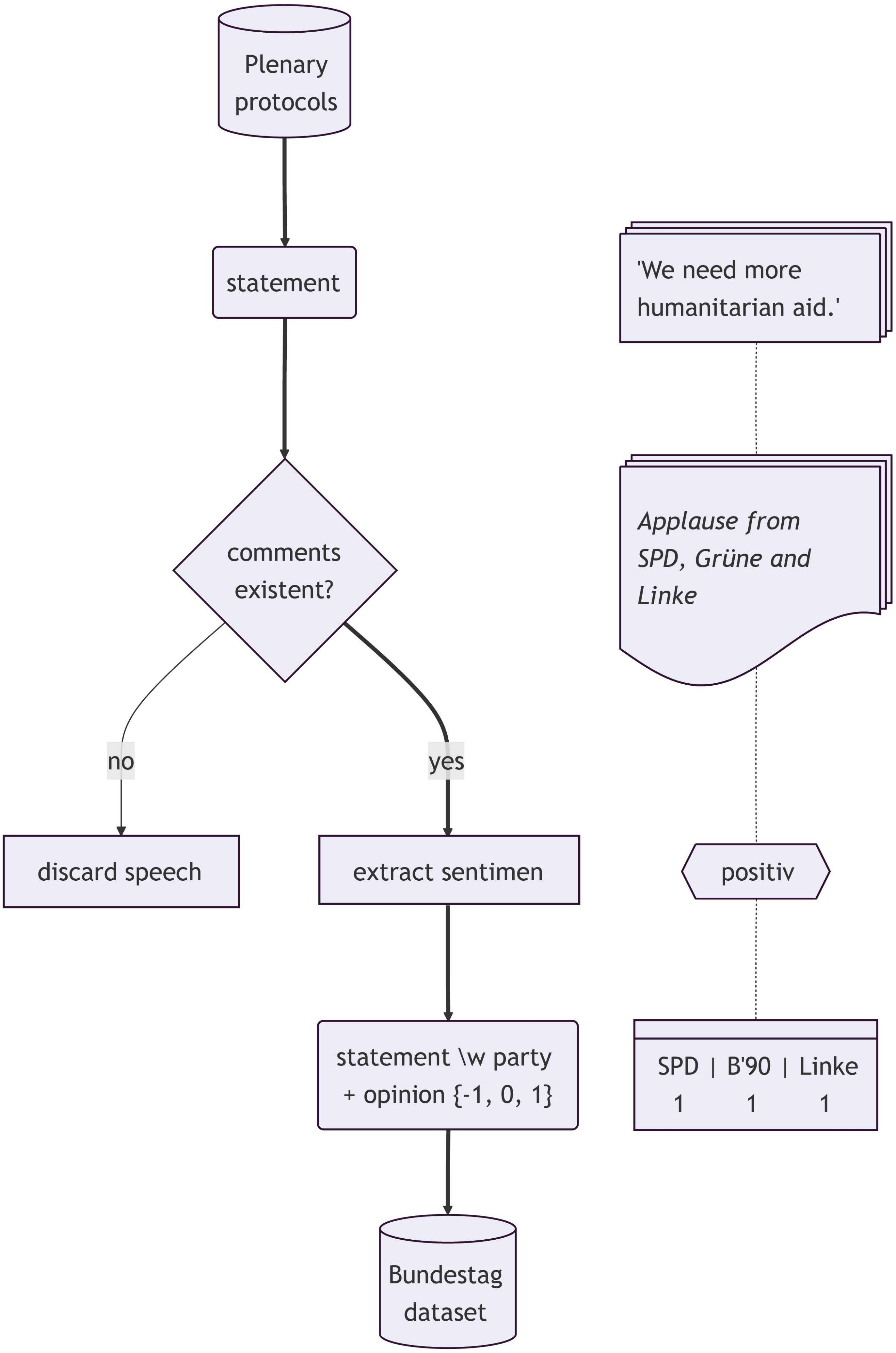}
  \caption{Flowchart of sentiment extraction}
  \label{fig:sentiment-bundestag}
\end{figure}

\begin{figure*}[t]
\centering
\includegraphics[width=1\textwidth]{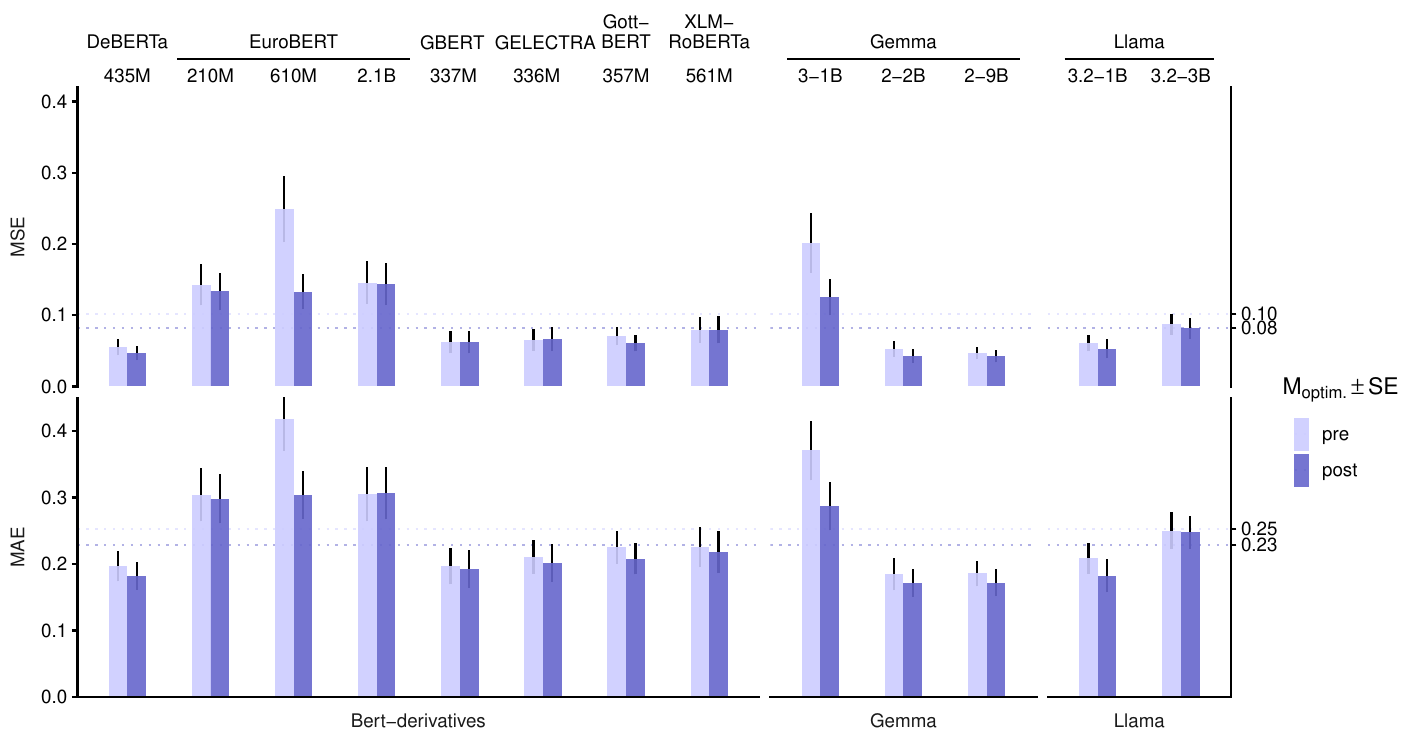}
\caption{Depicted is the effect of optimization across all 13 models and 33 news media outlets, measured using the mean absolute error (MAE, lower panel) and mean squared error (MSE, upper panel). Error bars represent standard errors (SE). Values are sorted by model class and, within each class, by parameter size for various Gemma, Llama, and Bert derivatives. The color contrast highlights the effect of the optimization: values before optimization (light bars) are generally higher than those after optimization (dark bars). This reduction in error metrics due to the optimization is evident from the dashed horizontal lines, which represent the mean values across the models. The differences indicate moderately strong effects, which we report as $d_{av}$ according to \citet{cumming_effectsize} with 95\% confidence intervals (CI). Specifically, the optimization had an estimated effect of $d_{av}=0.37$, $CI_{95\%}[0.08, 0.66]$ as measured by the MAE, and an only slightly smaller $d_{av}=0.36$, $CI_{95\%}[0.00, 0.73]$ with respect to the MSE. These effects were largely consistent across models and metrics, as reflected in high pre-post correlations ($r_\mathrm{MAE} = .91$ and $r_\mathrm{MSE} = .88$). Two exceptions stand out: EuroBERT-610M and Gemma-3-1B, for which optimization had a stronger effect regardless of the metric considered. These are also the models whose initial values were clearly above the average values (cf. the upper dashed line in both panels). No clear effect of model size (in terms of the number of parameters) on performance is evident; for both metrics and measurement points, size and error correlated only weakly with $r\approx-.25$ (for all metrics before and after optimization, with only post-optimization MSE showing a slightly higher correlation of $r=-.27$). In other words, a higher number of parameters tends to produce smaller errors across models, though this does not necessarily hold true for individual models. For example, the smaller Llama-3.2 with 1B parameters consistently yields lower errors than the much larger model with 3B parameters. The results suggest that model size alone is not a reliable predictor. At this point, it should be noted that these findings are reported purely descriptively; our setup did not have the primary goal of demonstrating an effect of model size but rather aimed to identify the best model. Here, Gemma2-2B yielded the lowest errors, regardless of the optimization or metric. However, the error bars suggest that the performance of much smaller models such as GBERT-337 or DeBERTa-425M does not differ significantly. No pairwise tests were calculated.
}
\label{fig:results-errorbars}
\end{figure*}

\end{document}